\documentclass[conference]{IEEEtran}

\usepackage{graphicx} 
\usepackage{subfigure} 

\usepackage{algorithm}
\usepackage{algorithmic}
\usepackage{amsmath, amsfonts}
\usepackage{amsthm}
\usepackage{comment}

\usepackage{hyperref}

\usepackage{csquotes}

\newtheorem{hypothesis}{Hypothesis}

\newtheorem{proposition}{Proposition}[section]

\newtheorem{definition}[proposition]{Definition}
\newtheorem{lemma}[proposition]{Lemma}

\newtheorem{theorem}{Theorem}
\newcommand{\bpf} {\noindent{\sc Proof} : }
\newcommand{\epf} {\hfill$\square$}
\newcommand{\E} {\mathbb{E}}

\renewcommand{\P} {\mathbb{P}}

\newcommand{\N}{\mathbb{N}}
\newcommand{\R}{\mathbb{R}}

\newcommand{\Z}{\mathbb{Z}}
\newcommand{\h}{\mathcal{H}}

\newcommand {\X} {\mathcal{X}}
\renewcommand {\O} {\mathcal{O}}
\newcommand {\y} {\mathcal{Y}}

\newcommand {\zi} {{Z^i}}

\newcommand{\Zs}{\mathcal{Z}}
\renewcommand{\L}{\mathcal{L}}

\DeclareMathOperator*{\argmin}{arg\,min}

\title{$m$-Power Regularized Least Squares Regression}

\author{\IEEEauthorblockN{Julien Audiffren}
\IEEEauthorblockA{Centre de Mathematiques et Leurs Applications\\
ENS Cachan, CNRS, Universit\'e Paris-Saclay,\\
94235, Cachan, France.}
\and
\IEEEauthorblockN{Hachem Kadri}
\IEEEauthorblockA{Laboratoire d'Informatique Fondamentale\\ Aix Marseille Univ, CNRS, LIF \\ Marseille, France}}

\begin{document}

\maketitle

\begin{abstract}
Regularization is used to find a solution that both fits the data and is sufficiently smooth, and thereby is very effective for designing and refining learning algorithms. But the influence of its exponent remains poorly understood. In particular, it is unclear how the exponent of the reproducing kernel Hilbert space~(RKHS) regularization term affects the accuracy and the efficiency of kernel-based learning algorithms. Here we consider regularized least squares regression~(RLSR) with an RKHS regularization raised to the power of $m$, where $m$ is a variable  real exponent. We design an efficient algorithm for solving the associated minimization problem, we provide a theoretical analysis of its stability, and we compare it with respect to computational complexity and prediction accuracy to the classical kernel ridge regression algorithm where the regularization exponent $m$ is fixed at $2$. Our results show that the $m$-power RLSR problem can be solved efficiently, and support the suggestion that one can use a regularization term that grows significantly slower than the standard quadratic growth in the RKHS norm.
\end{abstract}

\section{Introduction}

Regularization is extensively used in learning algorithms. It provides a principled way of addressing the well-known overfitting problem by learning a function that balances 
 fit and smoothness. The idea of regularization is hardly new. It goes back at least to~\cite{TIK63}, where it is used for solving ill-posed inverse problems.  Recently, there has been substantial work put forth to develop regularized learning models and significant progress has been made. 
Various regularization terms have been suggested, and different regularization strategies have been proposed to derive efficient learning algorithms. Among these algorithms one can cite regularized kernel methods which are based on a regularization over reproducing kernel Hilbert spaces (RKHSs)~\cite{SCH02,shawetaylor04}. 

A considerable amount of flexibility for fitting data is gained with kernel-based learning, as linear methods are replaced with nonlinear ones by representing the data points in high dimensional spaces of features, specifically RKHSs.
Many learning algorithms based on kernel methods and RKHS regularization~\cite{SCH02}, including support vector machines (SVM) and regularized least squares (RLS), have been used with considerable success in a wide range of supervised learning tasks, such as regression and classification. However, these algorithms are,  for the most part, restricted to a RKHS regularization term with an exponent equal to two. While the regularization hyperparameter has been extensively studied~(see e.g.~\cite{oneto2015support}), the influence of this exponent on the performance of kernel machines remains poorly understood. Studying the effects of varying the exponent of the RKHS regularization on the regularization process and the underlying learning algorithm is the main goal of this research.

To the best of our knowledge, the most directly related work to this paper is that of Mendelson and Neeman~\cite{Mendelson10} and Steinwart et al.~\cite{Steinwart09}, who studied the impact of the regularization exponent on RLS regression (RLSR) methods from a theoretical point of view. In~\cite{Mendelson10} the sharpest known learning rates of the RLSR algorithm was established in the case where the exponent of the regularization term $m$ is less than or equal to one, showing that one can use a regularization term that grows slower than the standard quadratic growth in the RKHS norm. 
%
%
However, in~\cite{Steinwart09} optimal learning rates independent of the exponent of the RKHS regularization was provided for the same algorithm when $m\geq 1$, arguing that the exponent $m$ has no influence on the learning rates and thus may be chosen on the basis of algorithmic considerations.
In this spirit we have asked whether, by additionally focusing attention on the algorithmic problem involved in the optimization, one could develop an efficient algorithm for RLSR with a variable RKHS regularization exponent. 
The remainder of the paper is devoted to presenting an approach to answering this question in the case of least square loss. %

It is worth mentioning that this question was asked in~\cite{Steinwart2004,Steinwart09} as an open-ended question. Indeed, even though Steinwart et al. showed that the same learning rates for RLSR can be achieved for regularization exponents greater or equal to 1,  they observed that a substantial difference is the way the regularization parameter has to be chosen in order to achieve this rate~\cite{Steinwart2004,Steinwart09}. 
This led them to wonder whether it is easier to find an almost optimal choice of the regularization parameter when the $m$-power regularization is considered instead of the standard quadratic one. This further motivates the study of the $m$-power RLSR problem from an algorithmic and impelmentation point of view.

{
In this work we demonstrate that the $m$-power RLS regression problem can also be solved efficiently and that there is no reason for ignoring this possibility. Specifically, we make the following contributions: 
\begin{itemize}
\parskip0pt \parsep0pt \itemsep5pt \topsep=0pt  \partopsep0pt

\item we derive a semi-analytic expression of the solution of the regularized least squares regression problem when the RKHS regularization is raised to the power of $m$, where $m$ is a variable real exponent,

\item we design a learning algorithm, $m$-RLSR, that computes the solution of the $m$-power RLS regression problem and able to achieve near-optimal performance without the use of cross-validation,



\item we provide a simple and efficient implementation of the proposed algorithm and enhance its scalability using random feature approximations of the kernel function,



\item we establish a theoretical result indicating that  the \mbox{$m$-RLSR} algorithm is uniformly stable when $m \geq 1$,

\item we experimentally evaluate the proposed algorithm and compare it to KRR with respect to prediction accuracy and optimal parameter search.

\end{itemize}

}

\section{Problem Overview}

This section presents the notation we use throughout the paper and introduces the problem of $m$-power regularized least squares regression.

\smallskip

\textbf{Notation. }
In the following, Let $m > 0$ be a real number, $\X$ a Hilbert space, $\h\subset \R^\X$ a separable reproducing kernel Hilbert space~(RKHS),  and $k : \X \times \X \rightarrow \R$ its  positive definite kernel. We suppose that $\sup_{x\in\X} k(x,x)=1$, which can be achieved by a proper rescaling as soon as $\sup_{x\in\X} k(x,x)<+ \infty$.
For all $n\in\N$,  Let $\Zs=\left\{ (x_1,y_1),..,(x_n,y_n)\right\} \subset \X \times \R$  denotes the training set constituted of $n$ realizations of the pair of random variable $(\mathbf{X,Y})$. We denote by $\P_\mathbf{X}$ the marginal law of $\mathbf{X}$.
Let $K$ be the Gram matrix associated to $k$ for $\Zs$  with~$(K_\Zs)_{i,j}= k(x_i,x_j)$. Finally, let $Y=(y_1,...,y_n)^\top$ be the output vector and $\vert \Zs \vert$ be the cardinal of the set $\Zs$. 
%

\smallskip

\textbf{M-power RLS regression. }
The algorithm we investigate here combines a least squares regression with an RKHS regularization term raised to the power of $m$. Formally, we would like to solve the following optimization problem:
\begin{equation}\label{eq Start}
\begin{aligned}
f_{Z,m,\lambda}=\argmin_{f \in \h} \frac{1}{n}\sum_{i=1}^n (y_i - f(x_i))^2 +  \lambda \|f\|^m_\h,
\end{aligned}
\end{equation}
where $m$ is a suitable chosen exponent.
Note that  the classical kernel ridge regression~(KRR) algorithm~\cite{saunders98} is recovered for $m=2$. 
The problem \eqref{eq Start} is well posed for $m>1$ as the function to minimize is strictly convex, coercive and continuous, hence it has a unique minimum. For $m<1$ the problem is no longer convex, but results on nonconvex optimization guaranty that under mild assumptions the \mbox{$m$-RLSR} with $m<1$ has a global minimizer, see e.g. \cite[Proposition 2.2]{bredies2009regularization}. 

%

\smallskip

 \textbf{KRR and $m$-RLSR: similar yet different.}
One crucial isue regarding the interpretation of the $m$-RLSR is whether by rescaling the regularization parameter, $m$-RLSR gives the same solution as KRR. Indeed, when $m >1$, the objective function of the $m$-RLSR optimization problem \eqref{eq Start} is strictly convex, and then by Lagrangian duality it is equivalent to its unconstrained version. In this case, it is possible to find a value of the regularization parameter such that the solution of the $m$-RLSR corresponds to that of the KRR. However, this is not the case when $m \leq 1$. This is summarized in the following Lemma.
%
%
\begin{lemma}\label{lemma weak equivalence}
$m$-RLSR and KRR optimization problem are equivalent in the following sense: $\forall \lambda >0,$ $m>1,$ $\Zs$ a training set, there exists $\lambda'>0$ such that $ \displaystyle f_{\Zs,m,\lambda}=f_{\Zs,2,\lambda'.}$
\end{lemma}

\begin{proof}
This results follows from the equivalence between Tikhonov and Ivanov regularization (see e.g. \cite{vasin1970relationship} and \cite[Chapter 5]{Rifkin02} and the fact that for $m>1$, \eqref{eq Start} is strictly convex.
\end{proof}
In other words, this lemma means that $m$-RLSR and KRR  share the same regularization path: $  \left\lbrace f_{\Zs,m,\lambda} \mathrel{}\middle|\mathrel{} \lambda \in \R_+ \right \rbrace =  \left\lbrace f_{\Zs,2,\lambda} \mathrel{}\middle|\mathrel{} \lambda \in \R_+ \right \rbrace$. 
However, this equivalence between optimization problems does not mean that the underlying learning algorithms are the same. Indeed, stochastic behavior and learning properties of those algorithms such as stability may greatly differ.  
In Section~4, we will study the stability of $m$-RLSR.
 Additionally, \cite{Steinwart09} have studied the generalization properties of $m$-RLSR and have shown that under some assumptions it achieves the same learning rate than KRR, but they observed that the regularization exponent may have an impact on the optimal choice of the regularization parameter. 
%
%

\smallskip

 \textbf{Optimal $\lambda$.} 
The $m$-RLSR problem have been studied only from a theoretical perspective (see e.g. \cite{Steinwart2005, Mendelson10,Steinwart09}). We recall here briefly some theoretical results of $m$-RLSR that are going to be used later in this paper, and we encourage the reader to refer to \cite{Steinwart09} for more details.
First, we need to define the integral operator associated to a reproducing kernel and two quantities, $p$ and $\beta$, which depends on $k$, $\mathbf{X}$ and $\mathbf{Y}$. 
\begin{definition}[Integral operator]
The integral operator associated to $k$ and $\mathbf{X}$ is defined as follows 
\begin{align*}
 T_k: \L_2(\P_\mathbf{X})& \rightarrow \L_2(\P_\mathbf{X}) \\
f(\cdot) &\mapsto \int_\X f(x) k(x,\cdot) d\P_\mathbf{X}(x),
\end{align*} 
where 
$$\L_2(\P_\mathbf{X})=\left\lbrace f : \X \rightarrow \R \quad \vert \quad  \int_{x\in \X} f(x)^2 d\P_\mathbf{X}(x) < \infty   \right\rbrace.$$

\end{definition}
It is well known (see e.g. \cite{steinwart2008support} Theorem 4.27) that $T_k$ is compact, so it has a countable number of non-zero eigenvalues. Let $(\mu_i)_{i \in \N}$ be these non-zero eigenvalues ordered in decreasing order.  We are interested in the rate of decrease of the sequence $(\mu_i)$, that is to say values of $p\in\left]0,1\right]$ such that 
\begin{equation}\label{eq def p}
\exists C>0\quad \forall i \in \N  \quad \mu_i < C i^{-1/p}. 
\end{equation}
Another quantity of interest is $\beta > 0$, which verifies
\begin{equation}\label{eq def beta}
\begin{aligned}
&\exists C>0 \quad \text{such that }\quad \forall \lambda>0 \\ 
&\inf_{f \in \h} \left( \lambda \|f \|_\h^2+ \| f(\cdot) - \E(\mathbf{Y} \vert \mathbf{X}=\cdot) \|_{\L_2(\P_\mathbf{X})}^2 \right)  \le C \lambda^\beta.\end{aligned}
\end{equation}

Now we recall the following result by Steinwart et al.~\cite[Corollary 6]{Steinwart09}:
\begin{proposition}\label{prop steinwart}
Let $p,\beta \in \left]0,1 \right[$ satisfying respectively \eqref{eq def p} and \eqref{eq def beta}.
If moreover $\mathbf{X}$ and $\mathbf{Y}$ are bounded a.s., then the \mbox{m-RLSR} problem with $m>1$ with the sequence of regularization parameters 

\begin{equation}\label{eq def lambda optimal}
\lambda_n= n^{-\frac{2\beta + m (1-\beta)}{2\beta + 2p }},
\end{equation}
where $n= | \Z | \rightarrow \infty$, achieves the asymptotically optimal learning rate $n^{-\beta/(\beta+p)}$, in the sense of it matches the lower bound given by \cite[Theorem 9]{Steinwart09}.
\end{proposition}
An immediate consequence of this result is that under the assumptions of Proposition~\ref{prop steinwart} the learning rate does not depends on $m$. 
However the optimal value of $\lambda_n$ cannot be computed in practice since it needs the knowledge of both $p$ and $\beta$. 
%
%
But in the particular case when $m=2p/(p+1)$, Equation \eqref{eq def lambda optimal} becomes

\begin{equation}\label{eq lambda for m<1}
\lambda_n=n^{-\frac{1}{1+p}},
\end{equation}
and no longer depends on $\beta$. Moreover, for typical RKHSs the value $p$ is known. This motivates the idea that $m$-RLSR  for specific values of $m$ might produce a \textquote{\textit{parameter-free}} learning algorithm~\cite{orabona2014simultaneous}. This  is discussed in Sex
ection 3 and verified in our experiments (see Section 5)


\section{$m$-Power Regularized Least Squares Regression Algorithm}\label{sect theory}
 \begin{table*}
    \normalsize
  \begin{center}
  \begin{tabular}{l}
    \hline \\[-1.0em]
    \textbf{Algorithm 1} \  \  \ $m$-Power RLS Regression Algorithm ($m$-RLSR) \hspace{0cm} \\
    \hline\\[-0.5em]
    \textbf{Input:}  training data $Z=\left\{ (x_1,y_1),..,(x_n,y_n)\right\}$, parameter $\lambda \in \mathbb{R^*_+}$, exponent $m \in \mathbb{R^*_+}$ \\[0.1cm]
    $\quad$ 1.\  \  \textbf{Kernel matrix:}  Compute the Gram matrix $K$ from the training set $Z$ :  $K = \big(k(x_i,x_j)\big)_{1\leq i,j\leq n}$  \\[0.2cm]
    $\quad$ 2.\  \  \textbf{Minimize the function} $\Gamma: \R^+ \mapsto \R$  defined by equation \eqref{eq def gamma} and obtain $\gamma$\\[0.2cm]
      $\quad$ 3.\  \  \textbf{Solution:}  Compute the solution $\alpha= (K +\gamma I)^{-1} Y$  \\[0.4cm]
    \hline
  \end{tabular}
    \end{center}
  \end{table*}

We now provide an efficient learning algorithm solving the $m$-power regularized least squares problem. 
It is worth recalling that the minimization problem \eqref{eq Start} with $m=2$ becomes a standard kernel ridge regression, which has an explicit analytic solution. 
In the same spirit, the main idea of our algorithm is to derive analytically from  \eqref{eq Start} 
a reduced one-dimensional problem on which we apply a minimization algorithm.

First, using the representer theorem~\cite{Dinuzzo12}, $f_Z$ can be written in the following form: 
$ f_Z= \sum_{i=1}^n \alpha_i k(.,x_i), $
with $\alpha_i \in \R$.
%
By combining this and~\eqref{eq Start}, the initial problem becomes
\begin{equation}\label{equation depart dual}
\begin{aligned}
\alpha=\argmin_{a \in \R^n} ( Y - K a ) ^\top ( Y - K a ) + n \lambda (a^\top K a )^{m/2},
\end{aligned}
\end{equation}
where $\alpha =(\alpha_i)_{1 \le i \le n}$ is the vector to determine. The following theorem gives an explicit formula for $\alpha$ that solves the optimization problem  \eqref{equation depart dual}.

\begin{theorem}\label{th algo}
Let $m>0$ and assume that \eqref{eq Start} has an unique solution. Then the function \begin{equation}\label{eq def gamma}
\begin{aligned}
\Gamma : \R_*& \rightarrow \R_+ \\
c& \mapsto c ^2 Y^T \left(K + c I \right)^{-2} Y \\
& \quad + n \lambda \left( Y^T \left(K + c I \right)^{-1} K \left(K + c I \right)^{-1} Y \right)^{m/2},
  \end{aligned}
\end{equation} 
has an unique minimum $\gamma$ and 
$$\alpha= (K + \gamma I)^{-1} Y, $$
is the unique solution of \eqref{equation depart dual}.
\end{theorem}

\begin{proof}
The proof is derived by setting the G\^{a}teaux derivative in $\alpha$ of the objective function \eqref{equation depart dual} equals to zero and using the obtained value of $\alpha$ in the original equation.
\end{proof}

It should be noted that $\Gamma$ is not convex in general, but \eqref{eq def gamma} can still be efficiently solved, as discussed in the algorithmic implementation below.

 \smallskip

 \textbf{Algorithmic Implementation of $m$-RLSR.} Theorem \ref{th algo} implies that the solution of the optimization problem \eqref{eq Start} is expressed analytically as a function	 of $\gamma$, the global minimum of $\Gamma$. Although $\Gamma$ is not convex in general, it is sufficiently smooth to be solved efficiently.  From a practical side, the minimization problem described in  \eqref{eq def gamma} has the advantage of being with respect with a single scalar value. From a general perspective, if $m>1$, the problem can be rewritten as finding the unique fixed point  of

$$F(c)\!=\!\frac{mn\lambda}{2} \left(\!Y^T ((K\!+\! cI)^{-1})^T\! K (K\! +\! c I)^{-1} Y \!\right)^{m/2 -1}.$$
In particular, it is interesting to note that for $m=2$, the previous equation immediately gives $\gamma=n\lambda$ and we retrieve the usual solution of the KRR.  Solving this problem is easier than the original because of smoothness of the function $F$ as long as $1-m/2 >0.5$, and a simple Newton's gradient descent can solve it quickly and efficiently.

The case $m<1$ should be dealt with other optimization tools.  As long as $m$ is not too close to zero, \eqref{eq def gamma} can be efficiently solved by using an adapted algorithm, such as a conjugate gradient descent using a dichotomous linear search and  Fletcher-Reeves criteria  (see e.g. \cite{Bredies2014,bredies2009regularization}). 
Our algorithm, which we call \mbox{\textit{$m$-power RLSR}}, uses these results to provide an efficient solution to regularized least squares regression with a variable regularization exponent~$m$~(see Algorithm 1).

\smallskip

 \textbf{About $p$ and $m$.} One of the interest of the $m$-RLSR is to take advantage of \eqref{eq lambda for m<1}. Indeed if $p$ is known, the (asymptotically) optimal value of the regularization parameter for the m-RLSR with $m=\frac{2p}{1+p}$ is also known, removing the need for a cross validation over a wide range of values to find  a good
regularization parameter. Before discussing the cases where $p$ is known, it is important to note that since $p\le 1$, so does $m$. As discussed above, the algorithm can be efficiently implemented even for $m<1$.

The quantity $p$ is known in some cases, such as {Sobolev space}.
Indeed If $\X \subset \R^d$ bounded, and $\P_\mathbf{X}$ is uniform distribution over $\X$, then  for all $q > d/2$,
the Sobolev space $W^q(\X)$ is a RKHS which satisfies \eqref{eq def p} with $p=d/2q$. 
Additionally 
the widely used RBF kernel satisfies \eqref{eq def p} for any $p>0$ if $\X$ is compact, (see e.g. \cite[Theorem 6.26]{steinwart2008support}).
We refer the reader to \cite{widom1964asymptotic} for more general results about $p$.

However when $m$ is close to zero, trying to solve \eqref{eq def gamma} directly produces numerical instability. Experimental results discussed in 
Section \ref{sect Xp} shows that taking an arbitrary value of $m$ more distant from to zero leads to good solution, and that the improvement obtained by taking a much lower but exact value (e.g $m=10^{-5}$) has to be put in balance with the computational complexity and precision issues due to the manipulation if those extreme values.
Also, it is interesting to note that as illustrated by our experiments, using approximate values for $p$ (and thus $m$) lead to a negligible decrease of accuracy. Even when $p$ is unknown, using the value of $\gamma$ obtained by solving \eqref{eq def gamma} with $\lambda= 1/n$ and $m=0.5$  leads to interesting results (although suboptimal).%

%
%

\smallskip

\textbf{Complexity analysis.} 
First we consider a naive implementation of the $m$-RLSR for input data of dimension $d$.
Gram matrix has complexity $\O(n^2d)$, while computing the solution  needs a matrix inversion which costs $\O(n^3)$. Then, the total complexity of a naive implementation of Algorithm 1 is $\O(n^3)$. Hence, naive $m$-RLSR achieves the same complexity as naive KRR. To improve the scalability of $m$-RLSR we use random features approximations of the kernel functions following the idea of [13]. The shift-invariant kernel function $k(x,x')$ in this case can be approximated by $g(x)^\top g(x')$ where $g$ is a mapping from $\R^d$ to $\R^D$ randomly drawn from
the Fourrier transform of the kernel function. $D$ is the dimension of the feature space and is very small compared to $n$ for large data sets. The complexity of the m-RLSR is reduced in this case to $O(ndD)$ which is the same complexity of the KRR with random features.

%


\section{Stability Analysis}\label{sect Stability}

\begin{table*}
\normalsize
\caption{Performance and running time of  $m$-power RLSR~(\mbox{$m$-RLSR}),  KRR, and their Random Feature(RF) counterparts algorithms on real dataset. The number of feature selected $D$ is mentioned in the last column.  }
\begin{tabular}{|l|l|l|l|l|l|l|l|l|l|}
\cline{2-10}
  \multicolumn{1}{c|}{}  & \multicolumn{2}{c|}{KRR} & \multicolumn{2}{c|}{$m$-RLSR}&\multicolumn{2}{c|}{KRR + RF} &\multicolumn{2}{c|}{$m$-RLSR + RF} &  \\
  \cline{1-1}
 Dataset & $\lambda=1/n$ & CV&  $m=0.5$ & $m=1$ & $\lambda=1/n$ & CV  &$m=0.5$ & $m=1$ & D\\
\hline

CPU  & 2\% &  1.6\%  & \textbf{0.9\%} & 1 \% &  4.3 \% & 3.8 \% & 3.2\% & \textbf{2.6 \%} & 300\\
  & 20 s & 4 mins & 21 s & 21 s &  8.2 s &  70 s & 8.4 s & 8.4 s & \\
\hline

Cadata & 0.8\%& \textbf{0.5}\%& \textbf{0.5} \%& \textbf{0.5} \%& 1.0 \% & 0.7 \% &  0.8 \% & \textbf{0.6  \%} & 100 \\
 & 15 mins & 3 h & 15 mins & 15 mins & 6s & 1.5 mins & 6.2 s & 6.2 s & \\
\hline

Census& 1 \%& \textbf{0.9} \% & \textbf{0.9} \% & \textbf{0.9} \% & 1.1 \%& 1 \% &\textbf{ 0.9 \%}  &1 \%  & 500\\
 & 20 mins & 4 h & 20 mins & 20 mins & 81 s & 15 mins & 82 s & 82 sec& \\
\hline
YearPredictionMSD\footnotemark & - & -& -& -& 3.5e-2 \% &  3.3e-2 & 3.3e-2 &3.3e-2& 300\\
  & - & - & - & - & 5 mins & 1 h & 5 mins & 5 mins & \\
\hline 
\end{tabular}
\label{tab_real}
\end{table*}

In this section we study the algorithmic stability of the \mbox{$m$-RLSR} algorithm. This notion reflects the behavior of a learning algorithm following a change of the training data, and was used successfully by~\cite{Bousquet} to derive bounds on the generalization error of  kernel-based learning algorithms. As discussed previously, the stability properties of KRR give no insight about the stability of $m$-RLSR, since the two algorithms are only weakly-equivalent. In this section we prove that \mbox{$m$-RLSR}  is stable for $m\ge1$.

In the following we denote by $\mathbf{X}$ and $\mathbf{Y}$ a pair of random variables following the unknown distribution~$P$ of the data, $\mathbf{X}$ representing the input and $\mathbf{Y}$ the output, by $\zi=Z\setminus(x_i,y_i)$ the training set from which was removed the element $i$. Let $c(y,f,x)= (y-f(x))^2$ denotes the cost function used in the algorithm. For all $f\in \h$, let $R_e(f,Z)=1/n \sum_{1\le i\le n} c(y_i,f,x_i)$ be the empirical error and $R_r(f,Z)=R_e(f,Z)+\lambda \|f\|^m_\h$ be the regularized error.
Let us recall the definition of uniform stability.

\begin{definition}[Uniform stability~\cite{Bousquet}] 
An algorithm $Z \rightarrow f_Z$ is said $\beta$ uniformly stable if and only if $\forall n \ge 1 $, $\forall 1 \le i \le n,$  $\forall Z $ a realization  of $n$ i.i.d. copies of $(\mathbf{X},\mathbf{Y})$,$ \forall (x,y) \in \X \times \y$ a $Z$ independent realization of $(\mathbf{X},\mathbf{Y})$,
 we have $\vert c(y,f_Z,x)- c(y,f_\zi,x)\vert \le \beta.$
\end{definition}

To prove the stability of a learning algorithm, it is common to make the following assumptions.

\begin{hypothesis}\label{Hyp Y bounded}
$\exists C_y>0 \text{ such that } \vert\mathbf{Y}\vert < C_y \text{ a.s.}$
\end{hypothesis}

\begin{hypothesis}\label{Hyp K bounded}
$\exists \kappa>0 \text{ such that }\sup_{x\in \X} k(x,x) < \kappa^2$.
\end{hypothesis}

The stability of our algorithm when $m\ge 1$ is established in the following theorem, whose proof uses the generalized Newton binomial theorem to extend the result of Theorem~22 in \cite{Bousquet}.

\begin{theorem}[Uniform stability of $m$-RLSR]\label{beta stable}
Let $C_y>0$ such that $\mathbf{Y} < C_y$ a.s and $\kappa>0$ such that $\sup_{x\in\X} k(x,x) < \kappa$, then the $m$-RLSR algorithm with regularization parameter $\lambda$ is $\beta$- uniformly stable $\forall m>1$ with 

$$ \beta=
\sigma \kappa  \left(    2^{m-1}\frac{\sigma \kappa  }{\lambda n b}\right )^{\frac{1}{m-1}}, $$
where $\sigma=2 \left(C_y +  \kappa\left(\frac{C_y ^2}{\lambda}\right)^\frac{1}{m}\right)$ and $b=2$ if $m>2$ else  $b=2^{m-1} \left( C_y^2 / \lambda\right) ^{1- 2/m}$.

\end{theorem}

The following Lemma is necessary to extend the original proof to case $m\ge 1$. 

\begin{lemma}\label{lemma c is lips}
If Hypotheses 1 and 2 hold, then $\forall n \ge 1 $, $\forall 1 \le i \le n,$  $\forall Z $ a realization  of $n$ i.i.d. copies of $(\mathbf{X},\mathbf{Y})$,$ \forall (x,y) \in \X \times \y$ a $Z$ independent realization of $(\mathbf{X},\mathbf{Y})$,
$$\vert c(y,f_Z,x)- c(y,f_\zi,x)\vert \le \sigma \vert f_Z(x) - f_\zi(x) \vert,$$
with $\sigma=2 \left(C_y +  \kappa\left(\frac{C_y ^2}{\lambda}\right)^\frac{1}{m}\right) $.
\end{lemma}

\bpf 
Since $\h$ is a vector space, $0 \in \h$, and 

\begin{equation}\nonumber
\begin{aligned}
  \lambda \|f_Z\|^m &\le\frac{1}{n}\sum_{i=1}^n (y_i - f_Z(x_i))^2 +  \lambda \|f_Z\|^m_\h \\
  &\le \frac{1}{n}\sum_{k=1}^n \|y_k-0 \|^2 +\lambda \|0\|^m_\h \\
  &\le C_y ^2, 
\end{aligned}
\end{equation}
where we used the definition of $f_Z$ and Hypothesis 2. 
Using the reproducing property and Hypothesis 3, we deduce that 
\begin{equation}\label{eq bound f_z(x)}
\nonumber
\begin{aligned}
\vert f_Z(x) \vert &\le \sqrt{k(x,x)} \|f_Z\|_\h \le \kappa\|f_Z\|_\h \\
& \le \kappa\left(\frac{C_y ^2}{\lambda}\right)^\frac{1}{m}. 
\end{aligned}
\end{equation}
The same reasoning holds for $f_\zi$. Finally,
\begin{equation}\nonumber
\begin{aligned}
\vert c(y, & f_Z,x)- c(y,f_\zi,x)\vert =\vert (y - f_Z(x))^2 -  (y - f_\zi(x))^2 \vert\\
&\le 2 \left(C_y +  \kappa\left(\frac{C_y ^2}{\lambda}\right)^\frac{1}{m}\right)  \vert f_Z(x)-f_\zi(x) \vert .      
\end{aligned}
\end{equation}      \epf
\begin{figure*}\label{figure variation}
\begin{center}
\includegraphics[scale=0.3]{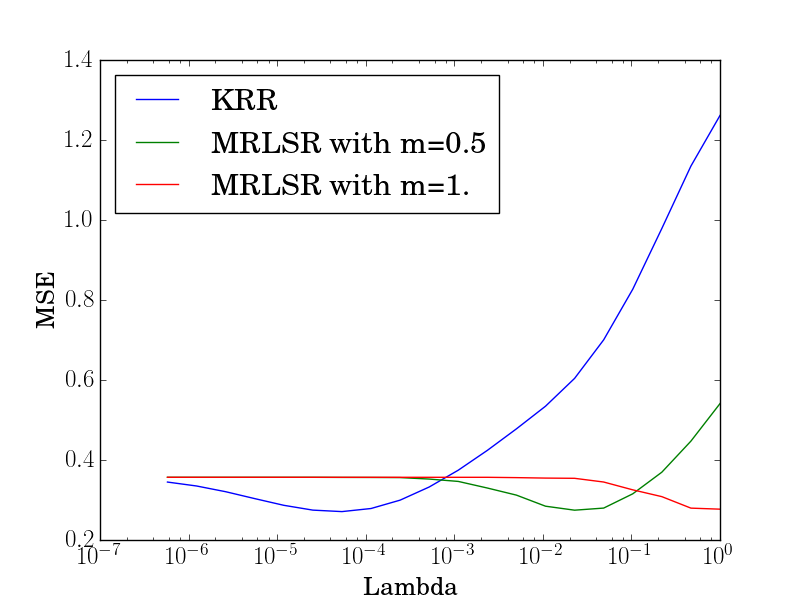} \hspace{-2em}
\includegraphics[scale=0.3]{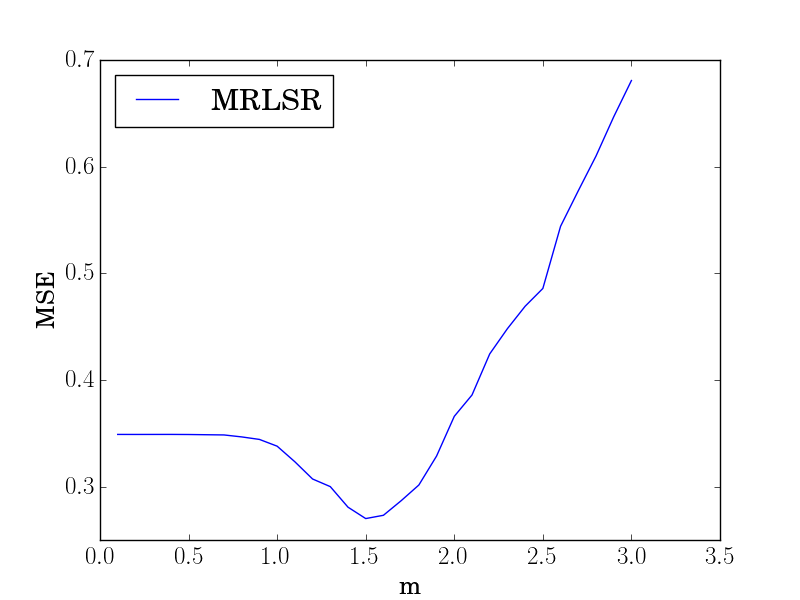}\hspace{-1.5em}
\includegraphics[scale=0.3]{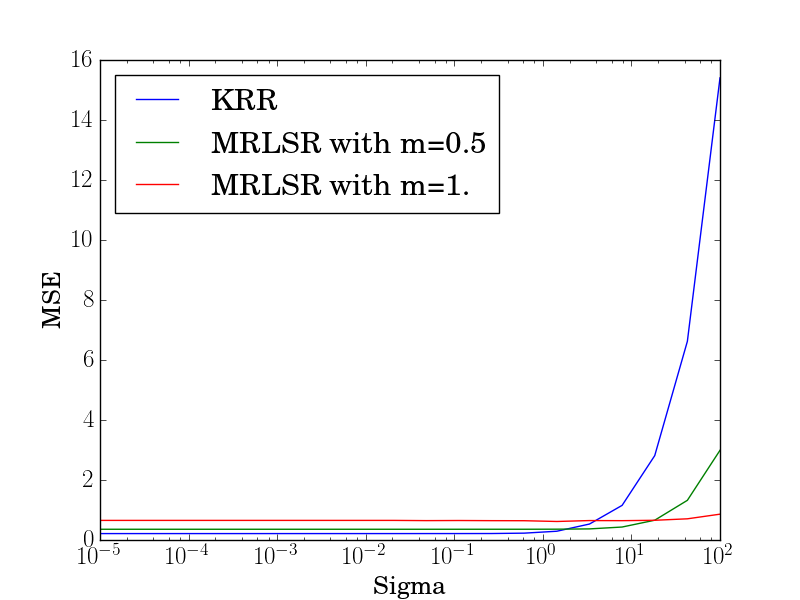}
\caption{Influence of the variation of the parameters on the $m$-RLSR. Left : $\lambda$ is varying, Middle : $m$ is varying, Right: $\sigma$ is varying}
\end{center}
\end{figure*}
\begin{proof}[Proof of Theorem \ref{beta stable}:]

By following the proof of Theorem 22 in \cite{Bousquet} with $t=1/2$, we obtain that
\begin{equation}\label{eq inequality with 1/2}
\begin{aligned}
\vert c(&y_i,f_Z,x_i) -c(y_i,f_Z + \frac{1}{2}(f_\zi-f_Z),x_i) \vert\\
& \ge n \lambda \left( \| f_Z \|^m_\h - 2 \left\| \frac{f_\zi+f_Z}{2}\right\|^m_\h
+ \| f_\zi \|^m_\h  \right),
\end{aligned}
\end{equation}
Let $u=(f_Z+f_\zi)/2$ and $v=(f_Z-f_\zi)/2$. Then,
\begin{equation}\nonumber
\begin{aligned}
&\| u +v \|^m_\h+ \|u-v \|^m_\h - 2 \left\| u\right\|^m_\h
- 2 \left\| v\right\|^m_\h\\
&\hspace{0.1cm}=\! \left( \|u\|^2_\h  \! +\! \|v\|^2_\h \! +\! 2 \left\langle u,v \right\rangle_\h \right)^{m/2}-\!2 \left( \| u\|^2_\h\right)^{m/2}\\
& \hspace{0.1cm} +\! \left( \|u\|^2_\h \!+\! \|v\|^2_\h \!-\! 2 \left\langle u,v \right\rangle_\h \right)^{m/2}\!\!
-\!2 \left( \| v\|^2_\h\right)^{m/2}\\
&\hspace{0.1cm}\ge 2\left( \|u\|^2_\h + \|v\|^2_\h \right)^{m/2} \!\!-2 \left( \| u\|^2_\h\right)^{m/2}
\!\!\!-2 \left( \| v\|^2_\h\right)^{m/2}\ge0,\\
\end{aligned}
\end{equation}
where in the last transition we used both Newton's generalized binomial theorem for the first inequality and the fact that $m/2 >1$ for the second one. Hence, we have shown that
\begin{equation}\label{eq inequality to prove}
\begin{aligned}
\| f_Z \|^m_\h - 2 \left\| \frac{f_\zi+f_Z}{2}\right\|^m_\h
+ \| f_\zi \|^m_\h \ge b \left\| \frac{f_\zi-f_Z}{2}\right\|^m_\h,
\end{aligned}
\end{equation}
with $b=2$.

For $2>m>1$, we have to proceed differently, and we obtain the same equality but with $b=2^{m-1} \left( C_y^2 / \lambda\right) ^{1- 2/m}$ by using \cite[Chapter~18, Theorem~3, p.~545 and p.518]{mpf}.A detailed proof can be found in the Appendix.

Now, by combining \eqref{eq inequality with 1/2} and \eqref{eq inequality to prove} and Lemma \ref{lemma c is lips}, the result follows.
\end{proof}

\footnotetext{Due to the size of the YearPredictionMSD dataset, we were unable to perform the KRR and the $m$-RLSR  algorithms without using the Random Features approximation. Also, the error are very small due to the nature of the problem and the definition of the error : the target is the year of the song, hence a one year error on each song will lead to a $\approx 0.05 \%$ error.}


\section{Experiments}\label{sect Xp}

In this section, we conduct experiments on synthetic and real-world datasets to evaluate 
the efficiency of the proposed algorithm.

\subsection{ Influence of $\sigma$, $m$ and $\lambda$}
\label{sec:xp_influence}


To see of the influence of the different parameters, we use a synthetic dataset (2000 instances, 10 attributes) described in \cite{tsang05}. In this dataset, inputs~$(x_1, . . . , x_{10})$ are generated independently and uniformly over $\left[ 0, 1\right]$ and outputs are computed from $y = 10 \sin(\pi x_1 x_2)+ 20(x_3 - 0.5)^2 + 10x_4 + 5x_5 + \mathcal{N}(0, \sigma).$ 

For these experiments we use a Gaussian kernel $k_\mu(x,x')=\exp(-\|x-x'\|_2^2  )$ and the relative
mean square error (RMSE), defined by $ \frac{100}{n}\sum_i (\frac{y_i - f(x_i) }{ y_i})^2 $ as the evaluation measure.

 We take particular interest in the value of $m \in \left]0,1 \right]$, as discussed in the previous sections.    We set $m=0.5$ or $1$, $\lambda=1/n$, $\sigma=2$, and we study the influence of those parameters on the algorithm.
 
 Figure 1  presents the evolution of the performance of the \mbox{$m$-RLSR} when the parameters $\lambda$, $m$ and $\sigma$ change.
It is interesting to note that \textbf{1)} the $m$-RLSR with $m\le 1$ seems to be less sensitive to the change of $\lambda$, and performs better if $\lambda$ is chosen close to $1/n$, even for $m = 0.5$ which is very different from the optimal value, \textbf{2)} the $m$-RLSR seems to adapt itself to change in the variance of the noise by rescaling its regularization (up to some extend), and finally \textbf{3)} the variations of the MSE occur smoothly as $m$ changes, which makes approximations of $m$ meaningful. 

\smallskip

\subsection{Accuracy and Running Time for large datasets}

For these experiments we use a Gaussian kernel $k_\mu(x,x')=\exp(-\|x-x'\|_2^2 / \mu)$
with $ \mu= \frac{1}{n^2}\sum_{i,j} \|x_i - x_j\|_2^ 2$ computed on the training set.

For a global assessment of the performance on real data, we used the following real world datasets which are publicly available:
\begin{itemize}
\item CPU (8192 instances, 8 dimensions) \footnote{\label{ref data 1}\url{https://www.csie.ntu.edu.tw/~cjlin/libsvmtools/datasets/regression.html}}
\item Census ( 22474, 14 dimensions)
\item Cadata (20640 instances, 8 dimensions) \textsuperscript{\ref{ref data 1}}
\item YearPresictionMSD (515345 instances, 90 dimensions) \footnote{\url{https://archive.ics.uci.edu/ml/datasets/YearPredictionMSD }}
\end{itemize}

In this setting little is known about the distribution of $\mathbf{X}$ and $\mathbf{Y}$. We choose to use the arbitrary values $1$ and $0.5$ for $m$ as well as $\lambda= 1/n$ for the $m$-RLSR, according to the previous observations.

Although the result concerning the cross-validation from \cite[Theorem 8]{Steinwart09} requires a number of value of lambda \textit{polynomial} in $n=|Z|$, this quickly becomes computationally intractable. We rather use grid of 10 possible $\lambda$ equally logarithmically space between $1/n^2$ and $1$.

All the results are reported in Table \ref{tab_real}.
They illustrate that the $m$-RLSR with $m=1$ and $m=0.5$ (and in general values of  $m \in \left] 0, 1 \right]$) can perform well even without cross-validation. Although  the KRR with an in-depth cross-validation can perform better, this cross validation becomes quickly too expensive and we think that the $m$-RLSR can offer an alternative to this problem, even without much information about the distributions of $(\mathbf{X},\mathbf{Y})$. Additionally, 
 using Random Features methods in the $m$-RLSR seems to be a fast and efficient approximation whose accuracy is very close to the one of the $m$-RLSR with a much lower running time.

\subsection{Prediction Accuracy using cross validation on $m$}

\begin{table}
\small
\begin{center}

\caption{Performance (RMSE and STD) of  $m$-power RLSR~(\mbox{M-RLSR}) and  KRR algorithms on synthetic and UCI datasets. $m$ is chosen by cross-validation on a grid ranging from 0.1 to 2.9 with a step-size of $0.1$. }

\begin{tabular}{|l|cc|ccc|}
\cline{2-6}
  \multicolumn{1}{c|}{}  & \multicolumn{2}{c|}{KRR} & \multicolumn{3}{c|}{M-RLSR} \\
 \hline
 Dataset & RMSE & STD & $m$ & RMSE & STD\\
\hline

Compressive&8.04e-2&3.00e-3&1.5&7.40e-2&3.67e-3\\

Slump&3.60e-2&5.62e-3&1.0&3.70e-2&6.49e-3\\

Yacht Hydro&0.165&1.13e-2&0.5&1.6e-2&7.53e-3\\

Wine&8.65e-2&6.18e-3&1.3&8.17e-2&6.07e-3\\

Energy&4.12e-2&1.79e-3&1.0&3.76e-2&2.87e-3\\

Housing&10.6e-2&7.98e-3&1.3&7.26e-2&9.92e-3\\

Parkinson&8.05e-2&4.51e-3&0.4&5.46e-2&3.29e-3\\

Synthetic&3.19e-2&1.56e-3&0.5&1.36e-2&5.85e-4\\

\hline

\end{tabular}
\label{tab_lambda}
\end{center}

\end{table}
This subsection aim to illustrate the influence of $m$ on the accuracy of the algorithm, as well as the the optimal values of $m$ chose by cross validation.

We use the following real-world datasets extracted from the UCI repository\footnote{ \url{http://archive.ics.uci.edu/ml/datasets}.}: Concrete Compressive Strength (1030 instances, 9 attributes), Concrete Slump Test (103 instances, 10 attributes), Yacht Hydrodynamics~(308 instances, 7 attributes), Wine Quality (4898 instances, 12 attributes), Energy Efficiency (768 instances, 8 attributes), Housing (506 instances, 14 attributes) and Parkinsons Telemonitoring (5875 instances, 26 attributes). Additionally, we also use the synthetic dataset described in Section~\ref{sec:xp_influence}.
%
In all our experiments, we use a Gaussian kernel $k_\mu(x,x')=\exp(-\|x-x'\|_2^2 / \mu)$
with $ \mu= \frac{1}{n^2}\sum_{i,j} \|x_i - x_j\|_2^ 2$, and the scaled root mean square error (RMSE), defined by $\frac{1}{\max y_i} \sqrt{\frac{1}{n}\sum_i (y_i - f(x_i))^2 }$, as evaluation measure.
For each dataset we proceed as follows:  the dataset is split randomly into two parts (70\% for training and  30\% for testing), we set $\lambda=1$, and we select $m$ using cross-validation in a grid varying from  $0.1$ to $2.9$ with a step-size of $0.1$. The value of $m$ with the least mean RMSE over ten run is selected.Then, with $m$ now fixed, $\lambda$ is chosen by a ten-fold cross validation in a logarithmic grid of $7$ values, ranging from $10^{-5}$ to $10^2$.
Likewise, $\lambda_2$ for KRR is chosen by 10-fold cross-validation on a larger logarithmic grid of 25 equally spaced values between $10^{-7}$ and $10^3$.

RMSE and standard deviation (STD) results for $m$-RLSR and KRR are reported in Table~\ref{tab_lambda}.
We show that the $m$-power RLSR algorithm is capable of achieving a good performance results when $m<2$.
Note that the difference between the performance of the two algorithms \mbox{$m$-RLSR} and KRR decreases  as the grid of $\lambda$ becomes larger --because of the equivalence discussed in Lemma \ref{lemma weak equivalence}-- but in practice the use of a large grid is limited by computational costs.

Moreover, with fixed hyper-parameters, $m$-RLSR and KRR run in roughly same amount of time. For example, with the same grid of lambda, KRR takes about 35 sec and M-RLSR 41 sec for the Parkinson data set. With the Synthetic data set, KRR takes 3.41 sec and M-RLSR 4.09 sec.

\section{Conclusion}\label{sect End}
 {
In this paper we proposed $m$-power regularized least squares regression (RLSR), a supervised regression algorithm based on a regularization raised to the power of $m$, where $m$ is with a variable real exponent.
From a theoretical perspective, we shed some light on the exact relation between the $m$-RLSR and the KRR, and showed that the $m$-RLSR is uniformly stable for all $m>1$.
Our experiments show that this algorithm is less dependant on the choice of the regularization parameter (and thus cross-validation) to achieve good performance in term of accuracy and running time, compared to the KRR.
Future work  might include the study of  the extension of those results to other kernel-based learning algorithms. 
}

\bibliographystyle{IEEEtran} 
\bibliography{biblioinfo}

\appendix

\section{Stability of m-RLSR }
In this section we will prove that the $m$-RLSR is uniformly stable for $2>m>1$. First let us recall two inequality theorems which will be used to prove the stability of the MRLSR algorithm.

\begin{theorem}{\cite[Chapter~18, Theorem~3, p.~545]{mpf}}
Let $\X$ be a pre-Hilbert space and let $x,y \in \mathcal{X}$. If $0\leq m\leq 2$ then
$$(\|x\| + \|y\|)^m + | \|x\| - \|y\| |^m \leq \|x+y\|^m + \|x-y\|^m  $$
and 
$$\|x+y\|^m + \|x-y\|^pm  \leq 2(\|x\|^2+\|y\|^2)^{m/2}$$

\end{theorem}

\begin{theorem}{\cite[p.~518]{mpf}}
If $a,b \in \mathbb{R}$, then
\begin{align*}
|a+b|^m \geq |a|^m + m |a|^{m-1} b(sgn\ a) + C(m) \frac{|b|^2}{|a|^{2-m}+|b|^{2-m}},
\end{align*}
where $1< m< 2$ and $C(m)>0$ is a constant which does not depend on $a$ et $b$.
\end{theorem}

To prove the stability of the MRLSR when $1<m<2$, we follow the same steps as in the proof of Theorem \ref{beta stable} for $m\geq 2$, using the following result:

\begin{proposition}
Let $\h$ be an RKHS and $u,v \in \h$ such that  $\|u\| < \kappa$ and $\|v\| < \kappa$. Then, for all $1<m<2$, we have
\begin{align}
\|u+v\|^m + \|u-v\|^m - 2 \|u\|^m \geq A(m) \|v\|^2,
\end{align}
where $A(m)>0$ is a constant which does not depend on $u$ et $v$.
\end{proposition}

\bpf \begin{align*}
&\|u+v\|^m + \|u-v\|^m - 2 \|u\|^m \\
&  \quad \geq (\|u\| + \|v\|)^m + | \|u\| - \|v\| |^m - 2 \|u\|^m \\
& \quad \geq \|u\|^m + m \|u\|^{m-1}\|v\| + C(m) \frac{\|v\|^2}{\|u\|^{2-m}+\|v\|^{2-m}} \\ 
& \quad +  \|u\|^m - m \|u\|^{m-1}\|v\| + C(m) \frac{\|v\|^2}{\|u\|^{2-m}+\|v\|^{2-m}} \\
& \quad - 2 \|u\|^m\\
& \geq 2C(m) \frac{\|v\|^2}{\|u\|^{2-m}+\|v\|^{2-m}}\\
& \geq A(m) \|v\|^2,
\end{align*}
where $A(m) = 2^{m-1} \kappa^{m-2}C(m)$. In the second line, we used Theorem 3, and in the third and fourth line, we used Theorem~4. \epf

We obtain that MRSLR with $1<m<2$ is $\beta$-stable with a stability of order of $O(n{^{-1}})$, where $n$ is the number of examples. Note that $\|u\|$ and $\|v\|$ are bounded since they are bounded by $\|f_Z\|$ and $\|f_{Z^{-i}}\|$, the solution of the regularized minimization problem with an RKHS norm constraint for $n$ and $n^{-i}$ examples, respectively ($\|f_Z\|$ and $\|f_{Z^{-i}}\|$ are bounded from above).

\end{document}